\documentclass[conference]{IEEEtran}
\usepackage{times}

\IEEEoverridecommandlockouts

\usepackage[numbers,sort]{natbib}
\usepackage{multicol}
\usepackage[bookmarks=true]{hyperref}

\usepackage[tbtags]{amsmath}
\usepackage{amsfonts}
\usepackage{amssymb}
\usepackage{amsthm}
\usepackage{amsmath}
\usepackage{mathrsfs}
\usepackage{xspace}
\usepackage{xparse}
\usepackage{mathtools}
\usepackage{multirow}
\usepackage{graphicx}
\usepackage{subcaption}
\usepackage{xcolor}
\usepackage{color}
\usepackage{booktabs}

\usepackage{url}

\usepackage[inline]{enumitem}
\usepackage{tabularx}

\begin{document}

\title{
Online Adaptation of Terrain-Aware Dynamics for Planning in Unstructured Environments
}

\author{
William~Ward,
Sarah~Etter,
Tyler~Ingebrand,
Christian~Ellis,
Adam~J.~Thorpe,
Ufuk~Topcu
\thanks{W. Ward, T. Ingebrand, C. Ellis, A. Thorpe, U. Topcu are with the Oden Institute for Computational Engineering \& Science, University of Texas at Austin. S. Etter is with the Department of Computer Science, University of Texas at Austin. C. Ellis is also with the DEVCOM Army Research Laboratory. 
  Email: {\tt wward@utexas.edu, etter@utexas.edu, tyleringebrand@utexas.edu, christian.ellis@austin.utexas.edu, adam.thorpe@autin.utexas.edu, utopcu@utexas.edu}.
}
\thanks{Corresponding author: A. Thorpe}
}

\maketitle

\begin{abstract}
Autonomous mobile robots operating in remote, unstructured environments must adapt to new, unpredictable terrains that can change rapidly during operation. In such scenarios, a critical challenge becomes estimating the robot's dynamics on changing terrain in order to enable reliable, accurate navigation and planning. We present a novel online adaptation approach for terrain-aware dynamics modeling and planning using function encoders. Our approach efficiently adapts to new terrains at runtime using limited online data without retraining or fine-tuning. By learning a set of neural network basis functions that span the robot dynamics on diverse terrains, we enable rapid online adaptation to new, unseen terrains and environments as a simple least-squares calculation. We demonstrate our approach for terrain adaptation in a Unity-based robotics simulator and show that the downstream controller has better empirical performance due to higher accuracy of the learned model. This leads to fewer collisions with obstacles while navigating in cluttered environments as compared to a neural ODE baseline. 
\end{abstract}

\IEEEpeerreviewmaketitle


\section{Introduction}
Rapid adaptation to unknown environments and terrain is critical for autonomous mobile robots. 
In off-road navigation, unpredictable terrain features such as rocky paths, forest floors, and wet fields can cause skidding, tripping, or immobilization, jeopardizing the robot's ability to reach its objective.
Autonomous ground vehicles must therefore dynamically adjust their behavior to terrain-specific conditions. This adaptation is challenging because terrain variations directly alter system dynamics. For example, tire response to acceleration depends on surface friction. 
As a result, accurate terrain-aware dynamics models are critical for control, particularly in planning-based control frameworks such as model predictive control (MPC) or model predictive path integral control (MPPI), which depend on accurate forward predictions of the robot’s dynamics \cite{williams2016aggressive, han2024modelpredictivecontrolaggressive}. 
Estimating the robot's dynamics online in uncertain environments is a particularly challenging problem because it typically must be performed without prior data or knowledge of the environment. 
This challenge is especially apparent for mobile robots operating in locations where human intervention is impractical or impossible, such as extraterrestrial exploration or hazardous disaster response. Consequently, learned models for dynamics prediction or estimation must be able to rapidly adapt to new environments and terrain at runtime.

\begin{figure}
    \centering
    \includegraphics[keepaspectratio,width=\columnwidth]{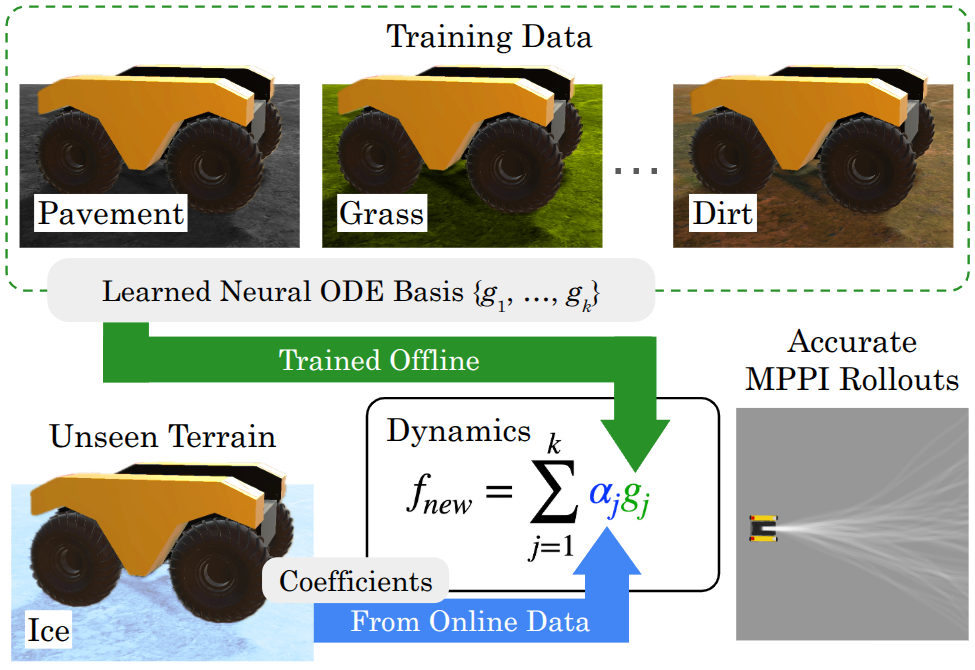}
    \caption{We rapidly identify the robot dynamics on a new, unseen terrain online using a set of learned neural ODE basis functions and a small amount of online data, leading to accurate, terrain-aware MPPI dynamics rollouts.}
    \label{fig: title figure}
\end{figure}

We present a method for modeling system dynamics that enables adaptation to new environments and conditions at runtime using limited online data. 
We focus on terrain adaptation for autonomous ground vehicles, though we note that our approach is broadly applicable to various systems and environment interactions.
Our approach uses the theory of function encoders, which have been applied in reinforcement learning \cite{ingebrand2024pmlr}, operator learning for PDE modeling \cite{INGEBRAND2025117646, thorpe2025basis}, and zero-shot dynamics estimation using neural ordinary differential equation (ODE) basis functions \cite{ingebrand2024FEnODEs}.
Rather than training a separate dynamics model for each terrain, we use function encoders to learn a set of neural network basis functions that spans a space of robot dynamics. 
At runtime, the dynamics are represented as a linear combination of the learned basis functions, thereby reducing the problem of dynamics estimation to that of identifying a vector of coefficients. We use this adapted model within a model-predictive controller to plan trajectories on unseen terrain, enabling closed-loop control without retraining.

State-of-the-art autonomous off-road navigation systems use a complex technology stack involving image segmentation \cite{viswanath2021offseg}, object identification, terrain estimation \cite{meng2023terrainnet, jung2024vstrong}, robot planning \cite{xiao2022motion}, and control \cite{10684236, 10756789, morrell2022nebula}. 
However, many approaches utilize simplified kinematic models, and neglect terrain-induced dynamics effects that are important for accurate MPC-based planning.
For instance, a ground robot equipped with a model tuned for driving on dry asphalt will experience significant tracking errors or loss of control when encountering slick, icy surfaces due to unmodeled changes in tire friction.

A common strategy for terrain-aware modeling is to classify the terrain using semantic segmentation or discrete labels \cite{wigness2019rugd, app10176044, pmlr-v164-shaban22a}, and then either switch to a corresponding dynamics model or adapt known models via parameter estimation. However, these approaches are only effective if the unknown terrain corresponds to a known model. In practice, this assumption fails in remote, unstructured environments, where terrain may be ambiguous or variable. 
For example, robots may encounter mixed terrain, such as patches of grass interspersed with mud, or micro-terrain features, such as loose gravel on top of hard soil, that does not fit neatly into predefined categories.

Our main contributions are: 
\begin{enumerate*}[label=(\arabic*)]
    \item 
    a method for terrain adaptation that combines function encoders, neural ODEs, and MPPI control, and 
    \item 
    an implementation of this approach in a realistic robotic simulator that demonstrates accurate closed-loop control on varied terrains.
\end{enumerate*}
While prior research has introduced function encoders for zero-shot prediction \cite{ingebrand2024FEnODEs}, we extend this approach to closed-loop control via MPPI. Specifically, we integrate the adapted dynamics model into an MPC pipeline to support real-time trajectory optimization on previously unseen terrain. The function encoder learns a set of neural ODE basis functions that span a space of vehicle dynamics, trained on trajectories from known terrains. In contrast to existing approaches, our approach collects limited data from a new terrain at runtime and rapidly adapts its coefficients online without retraining or fine-tuning. We implement and evaluate our approach in a high-fidelity Unity-based robotics simulator using a Clearpath Warthog robot. Figure \ref{fig: title figure} shows a visual representation of the proposed method.

\normalcolor
\subsection{Related Work}

\subsubsection*{Learning-based estimation and prediction}
Accurately predicting the dynamics of robotic systems operating in diverse, unstructured environments remains a critical yet challenging problem. Traditional approaches often rely on analytical models derived from first principles. However, the complexity and the difficulty of parameterizing real-world interactions, especially with varying terrain, significantly limit their applicability and accuracy in off-road settings.
Data-driven methods, leveraging neural networks to learn dynamics from experience, have shown promise in specific scenarios, including robotic manipulation \cite{byravan2017se3, gillespie2018learning}, autonomous driving on structured roads \cite{spielberg2019neural, nie2022deep}, and even highly dynamic maneuvers on controlled surfaces \cite{djeumou2023drift}. These learned models, however, typically struggle to generalize to novel terrains not encountered during training, which can lead to significant errors in trajectory prediction and control. The assumption of consistent ground conditions, inherent in many existing models, breaks down in off-road or unstructured environments where terrain properties change rapidly and unpredictably.

\subsubsection*{Transfer learning \& domain adaption}
Pre-trained dynamics models, while effective in their training domains, often exhibit poor performance when deployed in novel environments. Domain adaptation techniques aim to mitigate this by fine-tuning models using data collected in the new environment. In the context of off-road robotics, significant research has focused on capturing terrain-specific properties like friction, slip, deformability, and sinkage through parameter estimation \cite{pentzer2014online, li2018multi, padmanabhan2018estimation, espinoza2019vehicle, reina2020terrain, dallas2020online}. These methods typically involve robots estimating unknown terrain parameters from onboard sensor data at runtime. However, a critical limitation of this approach is the reliance on a predefined underlying dynamics model. If the true terrain dynamics deviate significantly from this assumed structure, parameter estimation will fail to provide accurate predictions.

Meta-learning approaches~\cite{finn2017model,xian2021hyperdynamics} expand these capabilities by training models specifically to adapt quickly to new tasks or environments with minimal data. These techniques relate closely to lifelong learning~\cite{liu2020learning, liu2021lifelong}, where systems continuously incorporate new knowledge while maintaining performance on previously learned tasks. Despite their advantages, meta-learning approaches ultimately still rely on some form of fine-tuning, which presents challenges in time-critical or resource-constrained scenarios. What is needed are models capable of online adaptation without retraining or fine-tuning. Our work addresses this by developing a dynamics estimation approach that adapts to novel terrains in real-time without explicit retraining. Related online policy adaptation methods use a two-stage training procedure to train a base policy and an adaptation module that predicts a latent vector that captures the effects of different terrains \cite{kumar2021rma}, such that the robot can adapt to changing terrains online without meta-learning.

Foundation models have demonstrated exceptional performance and adaptability across unstructured domains. In robotics, researchers have created diverse datasets \cite{o2023open} and developed generalizable models for various morphologies \cite{brohan2022rt, stone2023open, driess2023palm, team2024octo, black2024pi_0}, primarily using transformer architectures and leveraging behavior cloning. Yet they lack built-in mechanisms for adaptation. While our work focuses specifically on ground robot dynamics estimation across diverse terrains, we take a similar approach in principle and pre-train a function encoder model using diverse data before adapting predictions at runtime from observations.

\subsubsection*{Off-road driving at the limits}
Off-road autonomy presents unique challenges due to highly variable terrain conditions. Significant research has focused on estimating terrain traversability and hazard detection to identify safe paths \cite{Maturana-2017-102768, meng2023terrainnet, hdif2023}, yet comparatively less attention has been given to directly estimating dynamics and terrain interactions.
Terrain classification and semantic segmentation represent common adaptation approaches, where robots use sensor measurements to identify current terrain types \cite{ojeda2006terrain} and distinguish between traversable and non-traversable regions \cite{lalonde2006natural, mechergui2020efficient, castro2023feel, meng2023terrainnet, jung2024vstrong}. Typically, neural network classification and segmentation models are trained offline using data from various environments, then used at runtime to generate cost maps that improve local planning decisions.
However, this approach has important limitations. First, terrain classification is only useful when the robot already knows the associated dynamics. Otherwise, the robot must approximate using the closest available model, potentially leading to dangerous situations like attempting to use a pavement model on ice. Second, classification and segmentation primarily inform cost functions rather than dynamics functions, helping robots avoid dangerous areas but not necessarily navigate them when unavoidable.

\subsubsection*{Planning and control for autonomous navigation}
Adaptation strategies for control systems typically operate through either cost function or dynamics model adjustments. Local planners such as MPPI control use cost functions to evaluate planned trajectories and improve future control decisions. Prior work demonstrates that terrain-informed cost functions help robots avoid dangerous terrain \cite{lalonde2006natural, castro2023feel, meng2023terrainnet, jung2024vstrong}. 
However, when robots must traverse challenging terrain, such as large patches of ice, they require terrain-informed dynamics that enable local planners to determine how to navigate such terrain, beyond simply identifying areas to avoid.
Dynamics-based adaptation directly models how the robot interacts with different terrains. This approach allows control systems to anticipate and compensate for terrain-specific behaviors such as wheel slip or reduced traction~\cite{djeumou2024one,djeumou2023learngeneralizeminutesdata}. While cost function adaptation primarily influences path selection, dynamics adaptation affects both planning and execution by providing more accurate predictions of how control inputs will translate to robot motion across varying terrain types. Our work focuses on dynamics adaptation as it provides a more complete solution for autonomous navigation in challenging environments.

\section{Preliminaries \& Problem Formulation}
We seek to compute an estimate of robot dynamics online without prior data or knowledge of the environment.  
This is particularly challenging for mobile robots navigating unstructured environments.  
Specifically, we seek a model that is compatible with the Model Predictive Path Integral (MPPI) control framework \cite{williams2017information}.  
MPPI is a sampling-based optimization algorithm that computes an approximate solution to a optimal control problem at each control step.  
Below, we summarize its formulation and refer the reader to \cite{williams2017information} for additional details.  

Let $w \in \mathcal{W}$ denote the world state, which parameterizes the terrain effects on the dynamics.
The discrete-time system dynamics are given by,
\begin{equation}
    \label{eqn: system dynamics}
    x_{t+1} = F^{w}(x_{t}, v_{t}), 
\end{equation}
where $x_{t} \in \mathbb{R}^{n}$ is the state of the system at time $t$ and $v_{t} \in \mathbb{R}^{m}$ is the control input. 
Note that the world state $w$ is unobserved, meaning the true dynamics $F^{w}$ are unknown.
We assume that the actual control inputs applied to the system 
$V = \lbrace v_{0}, v_{1}, \ldots, v_{T-1} \rbrace$, over the horizon $T \in \mathbb{N}$, are not directly controllable. Instead, they are affected by noise such that $v_{t} \sim \mathcal{N}(u_{t}, \Sigma)$. 
The goal of MPPI is to compute a mean control sequence $U = \lbrace u_{0}, u_{1}, \ldots, u_{T-1} \rbrace$. 

Given a terminal cost function $\phi(x_{T})$ and a running cost function $\mathcal{L}(x_{t}, u_{t})$, the discrete-time optimal control problem is defined as,
\begin{equation}
    \label{eq: opt control problem}
    \min_{U \in \mathcal{U}} \mathbb{E} \biggl[ \phi(x_{T}) + \sum_{t=0}^{T-1} \mathcal{L}(x_{t}, u_{t}) \biggr],
\end{equation}
where $\mathcal{U}$ is the set of admissible control sequences and the expectation is taken with respect to the distribution over the control inputs $V$. We presume that the running cost $\mathcal{L}(x_{t}, u_{t})$ can be split into a state-dependent cost term $c(x_{t})$ and a control cost that is a quadratic function of $u_{t}$,
\begin{equation}
    \mathcal{L}(x_{t}, u_{t}) = c(x_{t}) + \frac{\lambda}{2} (u_{t}^{\top} \Sigma^{-1} u_{t} + \beta_{t}^{\top} u_{t}). 
\end{equation}
From \cite{williams2017information}, the affine term $\beta$ allows the location of the minimum control cost to be shifted away from zero.

Using a learned estimate $\hat{F}^{w}$ of the dynamics $F^{w}$ in \eqref{eqn: system dynamics}, the control algorithm computes $r$ rollouts, or state trajectories. 
From a given initial condition $x_{0}$ and control sequence $V$, a rollout is the predicted state trajectory $\lbrace x_{1}, \ldots, x_{T} \rbrace$ such that $x_{t+1} = \hat{F}^{w}(x_{t}, v_{t})$.
The state cost of a rollout is given by,
\begin{equation}
    \label{eqn: state cost}
    S(V, x_{0}) = 
    \phi(x_{T}) + \sum_{t=0}^{T-1} c(x_{t}).
\end{equation}

Each rollout is generated by randomly sampling sequences of control perturbations $V_{1}, \ldots, V_{r}$, which are random draws from the distribution over $V$. 
Then, we compute the state costs $S(V_{i}, x_{0})$, $i = 1, \ldots, r$, for each rollout.

After we obtain the costs, we compute probability weights and update the control sequence $\lbrace u_{0}, \ldots, u_{T-1} \rbrace$ by taking a probability-weighted average followed by a Savitsky-Galoy smoothing filter.
We then send the first control input $u_{0}$ to the robot, and the remaining sequence is used to initialize the optimization algorithm at the subsequent time instant. See \cite{williams2017information} for more details.

Importantly, the system model in \eqref{eqn: system dynamics} is effectively a black box.  
In MPPI, the control updates and optimization procedure are decoupled from the rollout generation process.  
This decoupling allows learning-based models to serve as drop-in replacements for analytical models.  
However, the performance of MPPI is highly sensitive to model fidelity.  
Model predictive control---and MPPI in particular---relies heavily on accurate forward models to generate rollouts.  
When there is mismatch between the learned model and the true system, the optimizer evaluates candidate control sequences under incorrect assumptions about how the system will evolve.  
As a result, it may select actions that appear optimal under the model but lead to poor or unsafe behavior in the real system.  
These errors are especially problematic in off-road environments, where unmodeled nonlinearities, contact dynamics, and terrain effects introduce significant model mismatch.  
Moreover, because learned models are typically trained in fixed conditions, they often fail to generalize to new environments without mechanisms for online adaptation. 

We consider the problem of learning a model \( \hat{F}^{w} \) of the system dynamics in \eqref{eqn: system dynamics}.  
The key challenge in the off-road setting is that the world state \( w \) is unobserved.  
To adapt to changing conditions and terrains, we seek a learned model that is both terrain-aware and computationally efficient for model-predictive control, suitable for deployment on hardware.  
During training, we presume access to historical trajectories of the robot operating on diverse terrains.  
From the trajectory data, we construct a set of datasets \( \{D^{w_1}, \ldots, D^{w_\ell}\} \).  
Each dataset \( D^{w_i} \) consists of states and actions collected from a single dynamics function \( F^{w_i} \),  
\begin{equation}  
    \label{eqn: dataset}  
    D^{w_i} = \left\lbrace (x_j, v_j, x_{j+1}) \right\rbrace_{j=0}^{m},  
\end{equation}  
where \( x_{j+1} = F^{w_i}(x_j, v_j) \).  
Such a dataset may be available, for instance, from historical trajectories observed during robot operation.

\section{Method}

\begin{figure}
    \centering
    \includegraphics[keepaspectratio,width=\linewidth]{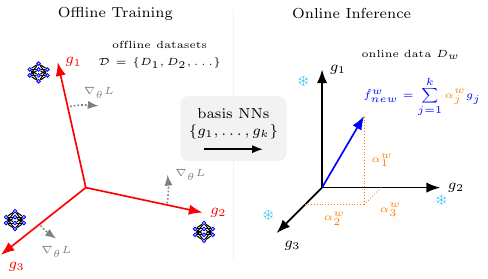}
    \caption{Function encoders consist of two steps: a training step to learn the neural network basis functions using offline datasets and an inference step using online data.}
    \label{fig: function encoder steps}
\end{figure}

We seek to learn a model of the system dynamics that can rapidly adapt to new terrains at runtime.
The evolution of the system can be expressed in terms of the equation,
\begin{equation}
    \label{eqn: discrete continuous model connection}
    x_{t+1} = F^{w}(x_{t}, v_{t}) = x_{t} + \int_{t}^{t+1} f^{w}(x(\tau), v_{t}) d\tau,
\end{equation}
where $f^{w}$ is the underlying vector field, and we integrate $f^{w}$ holding $v_{t}$ constant over the interval from $t$ to $t+1$. 

Neural ordinary differential equations \cite{chen2018neuralODEs} are a common approach for learning dynamical system models. 
These models approximate the vector field $f$ of a dynamical system with a neural network $f_{\theta}$.
Unlike discrete-time neural network models which map directly to the future states in a single forward pass, neural ODEs integrate over time in order to compute the solutions to initial value problems. 
Because neural ODEs compute the vector field of the system, they demonstrate excellent prediction accuracy without suffering from compounding errors that are inherent to discrete-time models.
Like standard Euler integration, discrete-time models accumulate prediction errors over long time horizons, which compound over time and can be detrimental for model predictive control that requires accurate rollouts.
However, neural ODEs do not have a built-in mechanism for online adaptation. 

To learn a model that can rapidly adapt to new terrains at runtime without retraining, we use function encoders with neural ODE basis functions \cite{ingebrand2024FEnODEs}.
Our work advances the existing function encoder approach by adapting the formulation with neural ODE basis functions to use least squares coefficient calculations and integrating it within MPPI control. 
In section \ref {section: results}, we demonstrate its effectiveness for ground robots navigating unstructured environments. 

\subsection{Learning Neural ODE Basis Functions}
Let $\mathcal{F} = \lbrace f^{w} \mid w \in \mathcal{W} \rbrace$ denote the space of all dynamics functions parameterized by the world state $w \in \mathcal{W}$.
The space $\mathcal{F}$ can be viewed as the space of all possible dynamics of the robot on different terrains, where the world state $w$ determines the terrain.
We assume that $\mathcal{F}$ is a Hilbert space. 
Note that this assumption is generally mild, since it imposes only mild regularity or smoothness conditions on the dynamics.
This assumption is key to our approach, since it allows us to learn a representation for $\mathcal{F}$.

We employ function encoders as in \cite{ingebrand2024FEnODEs} to estimate the dynamics. Figure \ref{fig: function encoder steps} shows a visual representation of function encoders. 
Function encoders learn a set of neural network basis functions $\lbrace g_{1}, \ldots, g_{k} \rbrace$ to span the Hilbert space $\mathcal{F}$. We can compute an estimate $\hat{f}^{w}$ of any function $f^{w} \in \mathcal{F}$ as a linear combination of the learned basis functions, 
\begin{equation}
    \label{eqn: linear combination}
    \hat{f}^{w} = \sum_{j=1}^{k} \alpha_{j}^{w} g_{j}
\end{equation}
where $\alpha^{w} \in \mathbb{R}^{k}$ are real coefficients corresponding to $f^{w}$. 
Intuitively, we can view the model as a linear combination of neural ODE basis functions,
\begin{align}
     x_{t+1} - x_{t} &= \sum_{j=1}^{k} \alpha_{j}^{w} \int_{t}^{t+1} g_{j}(x(\tau), v_{t}) d \tau \\
     \label{eqn: linear combination of neural odes}
     &= \sum_{j=1}^{k} \alpha_{j}^{w} G_{j}(x_{t}, v_{t})
\end{align}
where $G_{j}(x_{t}, v_{t}) := \int_{t}^{t + 1} g_{j}(x(\tau), v_{t}) d\tau$ is a neural ODE.
Note that the basis functions output the \emph{change} in state. This means that in practice, we generally need to account for this during integration by adding the state $x_{t}$ back at each integration step. 
The future state of the system $x_{t+1}$ can be computed as in \eqref{eqn: discrete continuous model connection} by integrating $\hat{f}^{w}$ as in \eqref{eqn: linear combination of neural odes}.

Importantly, we can quickly compute an estimate $\hat{f}^{w}$ of any dynamics function $f^{w} \in \mathcal{F}$ by identifying the coefficients $\alpha^{w} \in \mathbb{R}^{k}$.
The coefficients $\alpha^{w}$ can be computed efficiently in closed form as the solution to a least squares problem via the normal equation as, 
\begin{equation} \label{eqn: least squares}
\alpha^{w} = \begin{bmatrix}
\langle G_1, G_1 \rangle & \cdots & \langle G_1, G_k \rangle \\
\vdots & \ddots & \vdots \\
\langle G_k, G_1 \rangle & \cdots & \langle G_k, G_k \rangle \\
\end{bmatrix}^{-1}
\begin{bmatrix}
\langle F, G_1 \rangle \\
\vdots \\
\langle F, G_k \rangle \\
\end{bmatrix}.
\end{equation}
This requires us to compute the inverse of a $k \times k$ Gram matrix, which is generally $\mathcal{O}(k^{3})$. However, $k$ is generally chosen to be small, on the order of 100 or fewer, meaning the inverse can be computed efficiently. Additionally, we note that this computation can be performed once offline after training, and that we do not need to recompute the inverse during inference.
Since the basis functions output the change in state $x_{t+1} - x_{t}$, the inner product between functions $F$ and $G$ is defined as
\begin{equation}
    \langle F, G \rangle = \int_{\mathcal{X} \times \mathcal{U}} F(x, u)^\top G(x,u) d(x,u).
\end{equation}
We approximate the inner product using Monte Carlo integration \cite{ingebrand2024pmlr, ingebrand2024FEnODEs} using a small amount of input-output data of the form $(x_{t}, v_{t})$ and $x_{t+1} - x_{t}$ collected online.

\begin{figure*}[!h]
    \centering
    \includegraphics[width=\textwidth, height=2in]{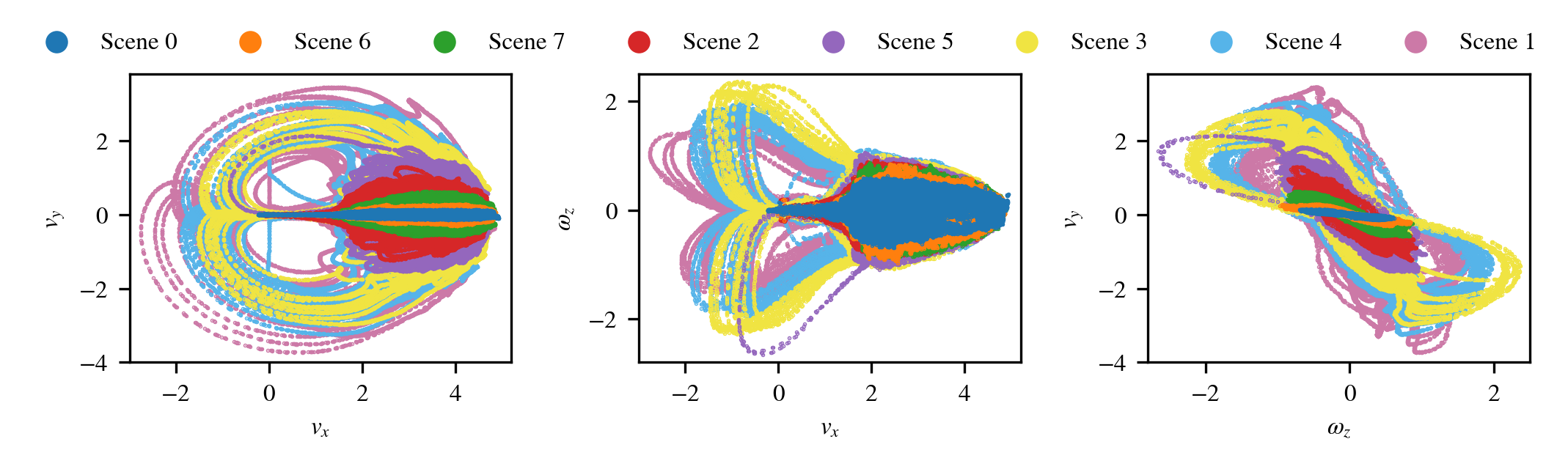}
    \caption{We collect data corresponding to diverse terrains that induce significant variation in the dynamics. Several terrains have low friction, which causes high lateral velocity.}
    \label{fig: data plots}
\end{figure*}

\subsubsection*{Offline Training}

For each function $f^{w_{i}}$ and the corresponding dataset $D^{w_{i}}$, we first compute the coefficients via \eqref{eqn: least squares} and then compute an empirical estimate $\hat{f}^{w_{i}}$ of $f^{w_{i}}$ as in \eqref{eqn: linear combination}. We then compute the MSE of the estimate using the norm induced by the inner product of $\mathcal{F}$. The loss is the sum of the errors for all $\hat{f}^{w_{i}}$, which is minimized via gradient descent. 
After training, the basis functions $\lbrace g_{1}, \ldots, g_{k} \rbrace$ are fixed. 
We denote the low-dimensional representation of $\mathcal{F}$ as 
\begin{equation}
    \label{eqn: learned space}
    \hat{\mathcal{F}} \coloneqq \overline{\mathrm{span}}\lbrace g_{1}, \ldots, g_{k} \rbrace.
\end{equation}

\subsubsection*{Online Inference}
At inference time, the basis functions do not require retraining or fine-tuning. For any new function $f^{w}_{new} \in \mathcal{F}$, we simply need to identify the coefficients of the dynamics representation $\hat{f}^{w}_{new}$ in the learned function space $\hat{\mathcal{F}}$ in order to estimate the function via \eqref{eqn: linear combination}. 
This representation is central to our approach, since it offers a means to quickly identify and estimate the dynamics. 
In practice, we typically require only a small amount of online data to compute $\alpha^{w}$. Generally, we can identify $\alpha^{w}$ using approximately 100 data points, which corresponds to a few seconds of online data.

As a remark, it is important to clarify the coefficients $\alpha^{w}$ correspond to a \emph{single} terrain. 
However, the terrain may change during off-road navigation, e.g.\ from dirt to sand to mud. 
We note that our proposed approach can likely be adapted to streaming data collected online via recursive least squares, which would offer a significant advantage over existing adaptation approaches such as meta-learning that require network fine-tuning, since we can use update our learned model via a simple coefficient update.

\subsection{Online Terrain Adaptation with MPPI}

At runtime, we collect a small sample of online data and solve the least squares problem \eqref{eqn: least squares} to estimate an approximate dynamics model $\hat{f}^w$ for $f^{w}$ as a linear combination of the learned neural ODE basis functions \eqref{eqn: linear combination of neural odes}. 
We then use $\hat{f}^w$ in place of the dynamics model in \eqref{eqn: system dynamics} to compute state trajectory rollouts from the robot’s current state and compute the control inputs from MPPI as normal.
Using the $r$ sampled control sequences $V_{1}, \ldots, V_{r}$, we generate predicted state trajectories using \eqref{eqn: linear combination of neural odes} by integrating the model over the planning horizon $T$. 
Then, for each rollout, we compute the state costs $S(V_{i}, x_{0})$ as in \eqref{eqn: state cost} and compute the MPPI probability weights as normal. 

By incorporating terrain-dependent dynamics, we obtain more accurate trajectory predictions and corresponding state costs \eqref{eqn: state cost}. 
As a result, MPPI generates more accurate trajectories that guide it toward better candidate solutions of the optimal control problem \eqref{eq: opt control problem}. 
Moreover, the rollouts for MPPI can be computed efficiently since they can be computed in batch, where each single-step prediction incurs only a cheap forward pass through $\hat{f}^w$. 
Most importantly, the function encoder model gives us terrain-aware predictions. 
Using function encoders with neural ODE basis functions to compute the rollouts for MPPI yields a computationally efficient online adaptation scheme that does not rely on expensive gradient updates or neural network fine-tuning.

\section{Results}
\label{section: results}

\begin{figure*}
    \centering
    \includegraphics[keepaspectratio, height=2.2in]{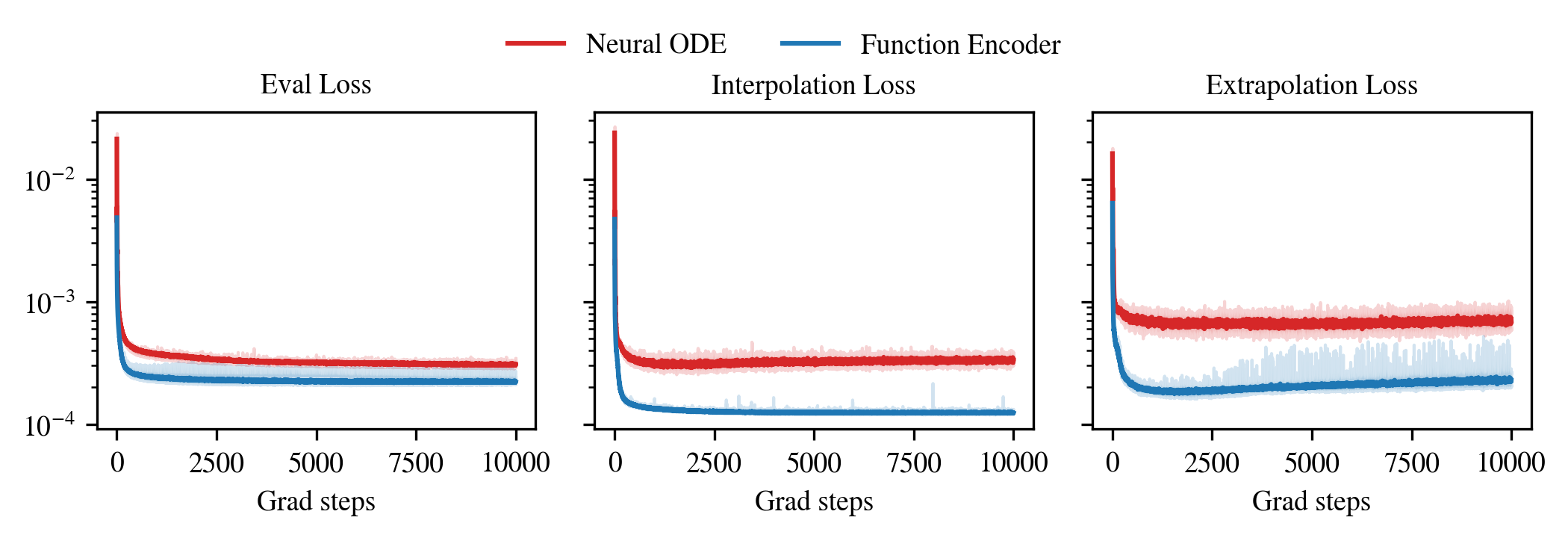}
    \caption{(Left) MSE of the function encoder versus the neural ODE model on unseen data from in-distribution scenes.  (Middle) MSE on an out-of-distribution scene whose dynamics is an interpolation of two in-distribution scenes. (Right) MSE on an out-of-distribution scene whose dynamics are more slippery than anything in the training sets, i.e., extrapolation. We evaluate over 10 random seeds. Each plot shows the median bounded by the minimum and maximum values (shaded regions).}
    \label{fig: training curves}
\end{figure*}

We demonstrate our proposed approach for terrain-aware planning and control in Phoenix, a Unity-based, high-fidelity robotic simulation environment designed for realistic outdoor navigation experiments.
Phoenix provides accurate dynamics modeling for robotic vehicles and supports obstacle-aware planning.
In this work, we evaluate our approach using a simulated Clearpath Warthog robot, a wheeled ground rover employing skid-steer drive.

The robot’s tires in Phoenix feature adjustable friction parameters, enabling simulation of various terrains such as pavement, mud, and ice.
Phoenix defines each tire’s forward and lateral friction behavior using five parameters: extremum slip, extremum value, asymptote slip, asymptote value, and stiffness \citep{unitydocs}.
Here, slip quantifies the relative rotation of the tire with respect to the ground before friction thresholds are reached, while value denotes the frictional force at these thresholds. 
The extremum represents the point of maximum frictional force, and the asymptote describes the steady-state friction force after surpassing the extremum slip.
Adjusting these parameters modulates the robot’s traction, effectively simulating diverse terrain interactions and driving dynamics.

We leverage the simulator’s adjustable friction parameters to replicate a variety of terrains. Specifically, we define two extreme parameter sets: a “normal friction” set representing pavement conditions, $a$, and a “low friction” set representing slick, icy conditions $b$. 
To systematically generate intermediate terrains, we select $6$ parameters in-between these two values $a$ and $b$ using a convex combination $\theta a + (1 - \theta) b$ of the parameter vector with $\theta \in [0, 0.25, 0.5, 0.75, 0.812, 0.875, 0.939, 1]$ to create a total of $8$ different terrains. 
Each resulting terrain corresponds uniquely to a different world state $w$.

\subsection{Dataset Collection and Pre-processing}
For each simulated terrain, we collect approximately 15 minutes of driving data using the simulator's integrated MPPI controller and analytical model guided by manually specified waypoints.
The dataset comprises odometry measurements and control commands sent to the robot.
Specifically, odometry data includes the robot's inertial frame coordinates $x$, $y$, heading angle $\psi$, linear velocities $v_x$, $v_y$, and angular velocity $\omega_z$, with velocities expressed in the robot's local (body) coordinate frame.
Additionally, we record the commanded forward and angular velocities throughout each run.

The collected dataset comprises measurements expressed in both inertial and body reference frames.
Because the system dynamics are invariant under translation and rotation, we apply a coordinate transformation to express the inertial states $(x, y, \psi)$ within the robot's local (body) frame.
Since the robot's dynamics remain consistent irrespective of absolute position or orientation, converting global states into the body frame significantly improves data efficiency and model learning.
In the inertial frame, odometry data collected from the simulator tends to be sparse and broadly distributed across a large range, leading to limited overlap among trajectories.
This sparsity can degrade model accuracy, motivating our transformation to the robot's local frame.
However, we acknowledge that such a coordinate transformation may not be feasible or practical in all real-world scenarios.

Figure~\ref{fig: data plots} shows the distribution of velocities recorded on each terrain.
On high-friction terrain, the robot exhibits minimal slip, and its $y$-velocity remains tightly bounded.
As terrain friction decreases, slip increases, particularly during acceleration.
In low-friction conditions, the robot frequently loses traction, resulting in significant lateral drift, spinouts, and occasional backward sliding.

From each terrain, we generate a dataset $D^{w_{i}}$ from the collected trajectory $\lbrace x_{0}, v_{0}, \ldots, x_{T-1}, v_{T-1}, x_{T} \rbrace$.
From the trajectory, we compute the time difference $\Delta t$ between consecutive states, as well as the state difference $\Delta x_{t} = x_{t+1} - x_{t}$.
We then construct a dataset of input-output pairs $(x_{t}, v_{t}, \Delta t, \Delta x_{t})$ for training the model as described in Equation~\eqref{eqn: linear combination of neural odes}.

\subsection{Model \& Implementation Details}

Both the neural ODE and function encoder (FE-NODE) models are trained on identical datasets.
Training was performed on a desktop equipped with an NVIDIA RTX 4060 GPU, Intel Core i7-14700 CPU, and 32GB RAM. 
The function encoder model required approximately $1$ hour and $15$ minutes of training, while the neural ODE model took approximately $21$ minutes.
For inference, both models used a Runge-Kutta 4 (RK4) numerical integrator to compute forward passes. 
In our implementation, the hyperparameter $k$ denoting the number of basis functions was set to $k=8$. The number of basis functions required depends on the system. However, this is generally unknown without prior knowledge. Therefore, we can overestimate with a larger number of basis functions than needed. For an ablation analysis of increasing the number of basis functions, see \cite{ingebrand2024FEnODEs, fe_transfer}.
The MPPI controller used in our experiments samples $1,000$ rollouts at each control iteration.
Each rollout was simulated over a time horizon of $10$ seconds with a sampling rate of $10$ Hz, resulting in predictions at $0.1$-second intervals.

The data processing pipeline for both models was identical.
As input, we use real-time odometry data, including linear and angular velocities, along with the commanded velocities.
Using this input, we compute forward passes through the respective learned dynamics models.
These predictions are then used to generate trajectory rollouts within the MPPI-based model predictive control algorithm.
Finally, control inputs are computed from the rollouts following the standard MPPI procedure.

\subsection{Training Results}
We plot the training curves in Figure \ref{fig: training curves}.  
The results indicate that the function encoder slightly outperforms neural ODEs on in-distribution scenes. 
However, for an out-of-distribution scene that is an interpolation of two of the training scenes, the neural ODE's performance degrades significantly. 
This is because the neural ODE is overfitting to the training scenes, whereas the function encoder adapts to this new scene via its coefficient calculation. This effect is even more pronounced for the held-out, extrapolation scene, where the neural ODE's performance suffers. Additionally, the training curve on the extrapolation dataset is indicative of overfitting for both algorithms. 

\subsection{Generalization and Online Adaptation Across Scenes}

\begin{figure}
    \centering
    \includegraphics[width=\columnwidth]{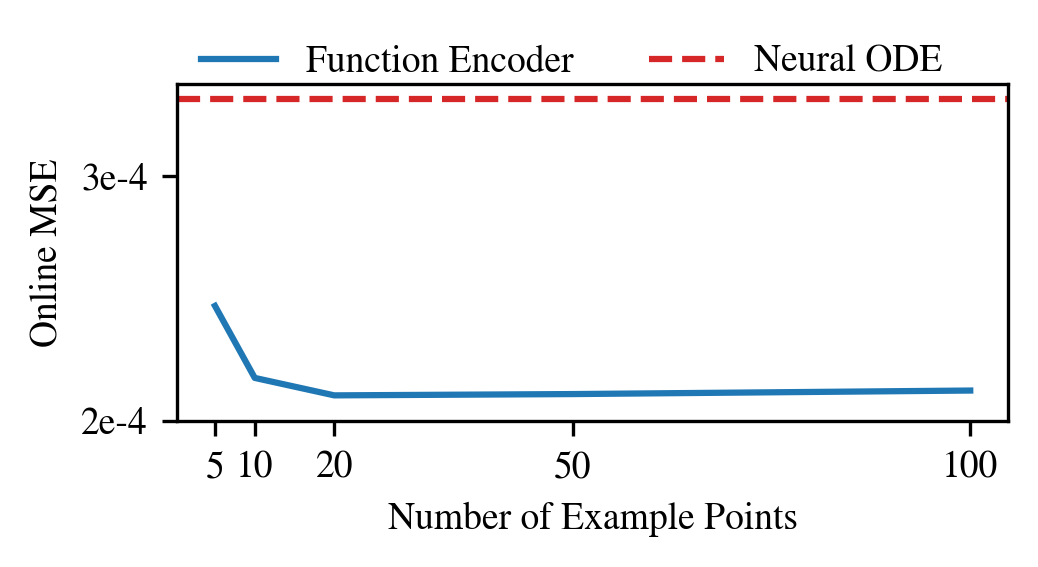}
    \caption{We evaluate the performance of the function encoder in an online setting. At each timestep, we use the most recent $n$ data points to compute the coefficients. This figure shows that the function encoder only requires a small amount of online data to get accurate 1-step predictions.}
    \label{fig: number of example points}
\end{figure}

To highlight our approach's ability to adapt to new settings, we mimic an online setting and measure the MSE of the next state predictions with only recent data to compute the coefficients. The results are shown in Figure \ref{fig: number of example points}. We iterate over the data for each scene, and at any given timestep, we use the previous $n$ data points to compute the coefficients. The resulting loss for the function encoder is the same as during training, suggesting that the function encoder is successfully adapting from the most recent $n$ data points. Furthermore, this curve ablates the amount of data needed to compute the coefficients. We find that 5 data points, which is approximately half a second, is enough for the function encoder to outperform neural ODEs. The function encoder's performance is stable as the amount of example data increases, suggesting that the coefficient estimate's accuracy has saturated. 

\begin{figure}
    \centering
    \includegraphics[keepaspectratio, width=\columnwidth]{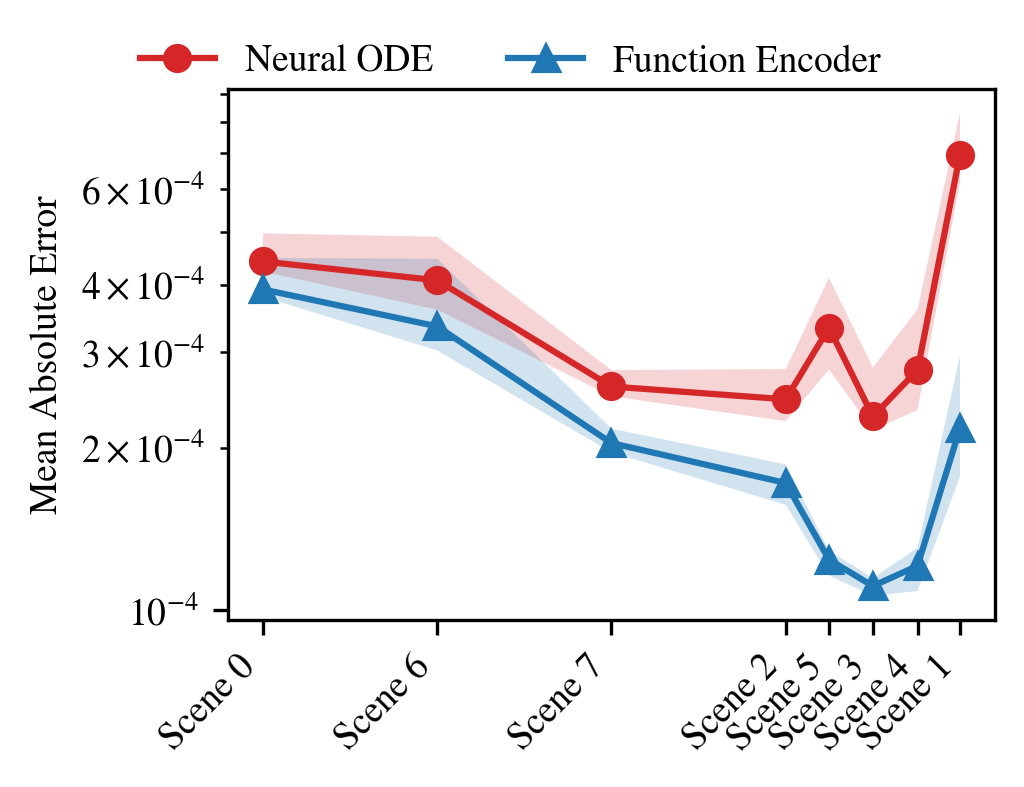}
    \caption{Error per-scene of our proposed approach compared to a standard neural ODE. Our approach achieves low error across all scenes, and does not suffer an increase in error for the interpolation dataset (scene 5). In contrast, neural ODEs perform much worse at this unseen, interpolation dataset. We evaluate over 10 random seeds. The median is bounded by the minimum and maximum values (shaded regions).}
    \label{fig: scene error comparison}
\end{figure}

In Figure \ref{fig: scene error comparison}, we plot the error per scene. We see that the function encoder slightly outperforms neural ODEs on all scenes, which corresponds with the training curves. 
Furthermore, we see that neural ODEs suffer an increase in error for scene 5, which is the held-out, interpolation scene. This is because neural ODEs are overfitting to the training scenes, whereas the function encoder uses a small dataset to calibrate to this new scene. Both approaches have worse performance on the extrapolation scene (Scene 1), although the function encoder outperforms neural ODEs on the validation set.

\begin{figure}
    \centering
    \includegraphics[keepaspectratio, width=\columnwidth]{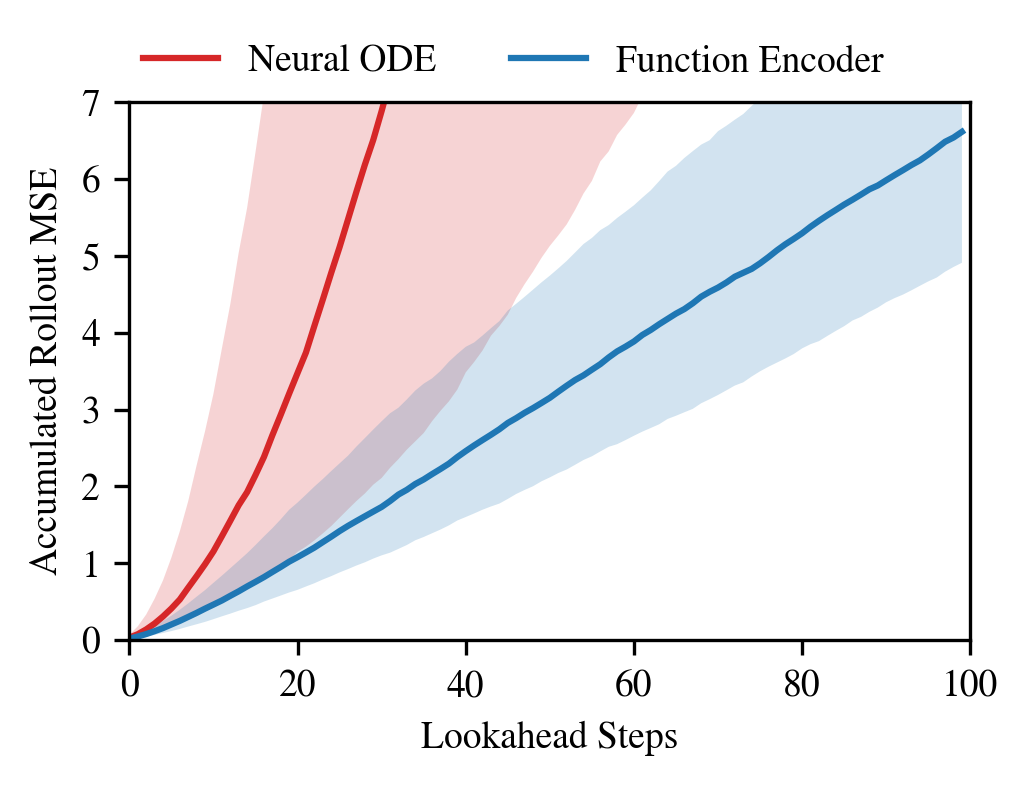}
    \caption{Accumulated MSE over time for scene 5. Our approach achieves more accurate rollouts over time on the interpolation scene (Scene 5). We evaluate over 10 random seeds and 100 rollouts per seed. The median error values are bounded by the 90\% and 10\% quartiles (shaded regions).}
    \label{fig:k-step-interp}
\end{figure}

\begin{figure}
    \centering
    \includegraphics[keepaspectratio, width=\columnwidth]{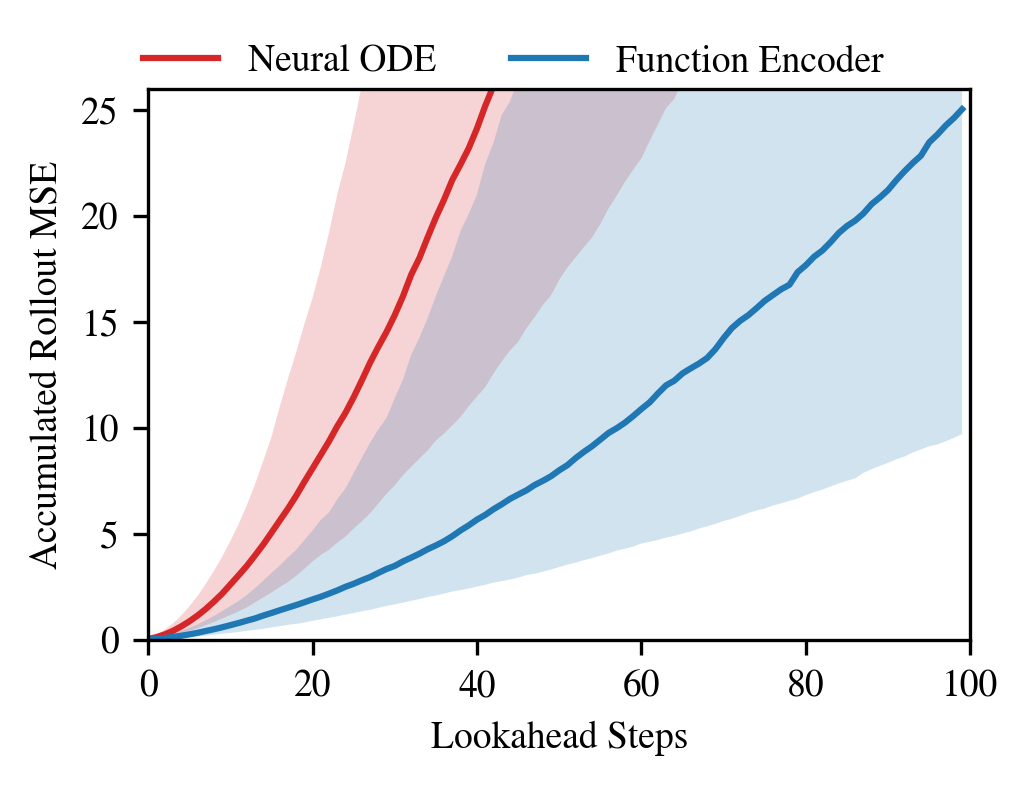}
    \caption{Accumulated MSE over time for Scene 1 (icy). }
    \label{fig:k-step-extrap}
\end{figure}

In Figures \ref{fig:k-step-interp} and \ref{fig:k-step-extrap}, we evaluate the accumulated MSE of each model's predictions over time. To simulate online trajectory rollouts as in MPPI, we choose random initial states from the scene data, and the models propagate the state forward using the recorded control inputs. Then, we evaluate the MSE between the true and predicted state after each prediction. Figure \ref{fig:k-step-interp} shows the results on the interpolation scene (Scene 5), and figure \ref{fig:k-step-extrap} shows the results on the extrapolation scene (Scene 1). In both plots, the neural ODE accumulates prediction error at a much faster rate than the function encoder. This indicates that the function encoder produces more accurate MPPI rollouts over time.

\subsection{Qualitative Results}

\begin{figure}
    \centering
    \includegraphics[width=\columnwidth]{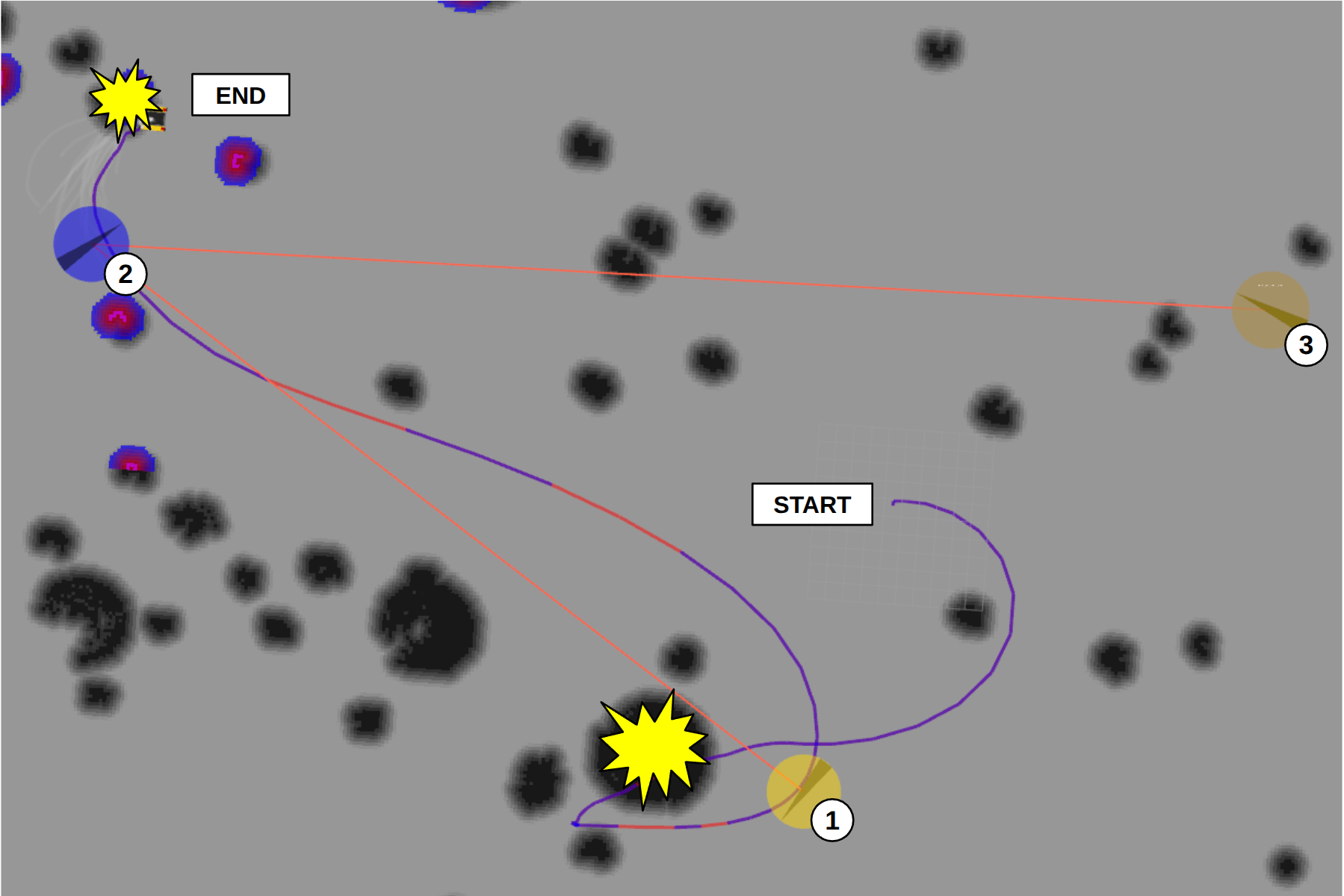}
    \caption{The neural ODE struggles to extrapolate to an unknown icy terrain. Due to poor rollout predictions, the robot collides with two trees and fails the mission.}
    \label{fig: phoenix extrapolation figure forest node}
\end{figure}

\begin{figure}
    \centering
    \includegraphics[width=\columnwidth]{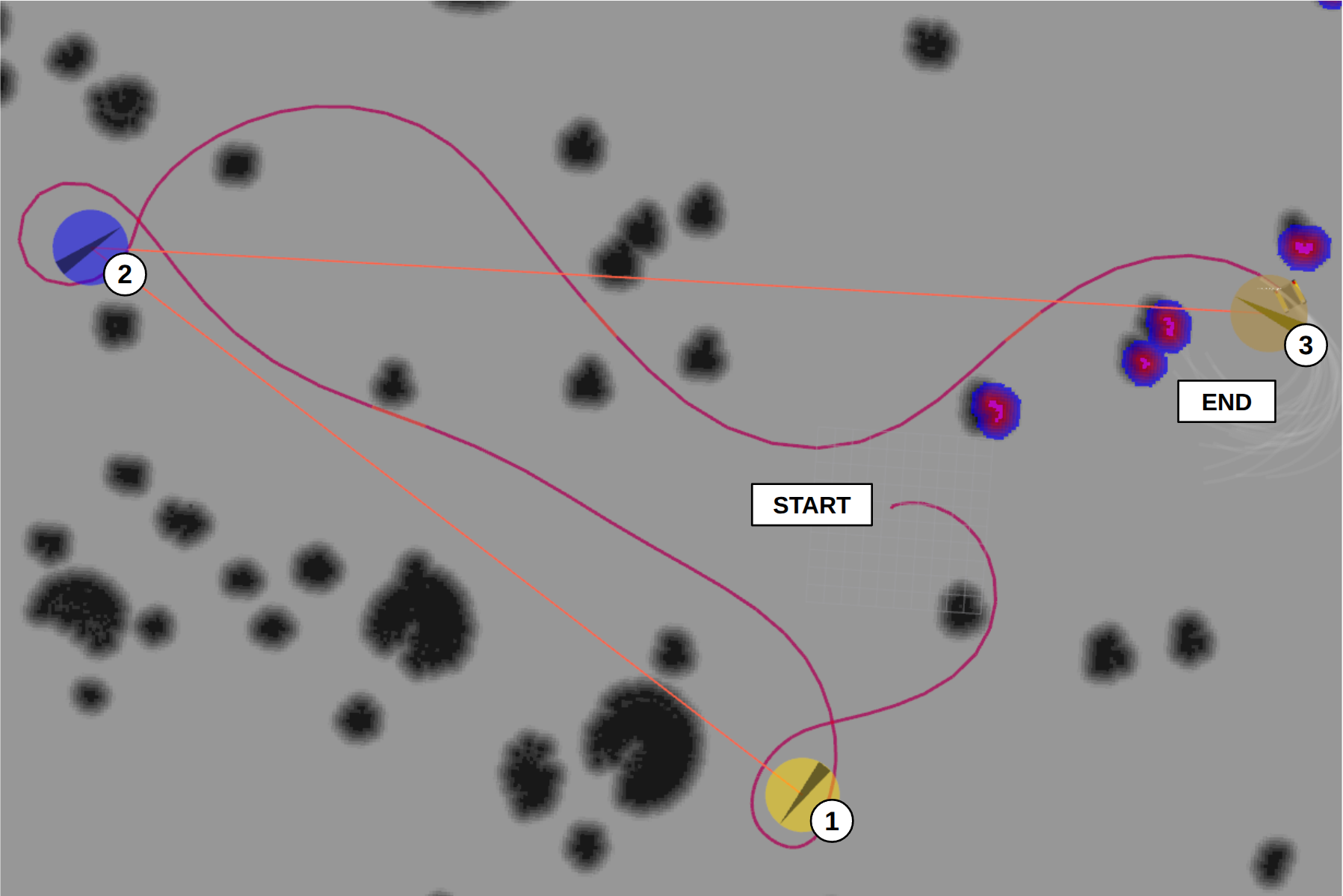}
    \caption{The function encoder successfully extrapolates to an unknown icy terrain, avoids all obstacles, and completes the mission.}
    \label{fig: phoenix extrapolation figure forest fe}
\end{figure}

We deploy the trained neural ODE and function encoder models on the simulated Warthog ground robot in Phoenix. The robot must navigate through a wooded area to reach a series of waypoints without colliding with trees. MPPI controls the robot during waypoint navigation using a learned model to calculate trajectory rollouts. Figures \ref{fig: phoenix extrapolation figure forest node} and \ref{fig: phoenix extrapolation figure forest fe} compare how the neural ODE and the function encoder perform in this navigation experiment while driving on the unknown ice terrain (Scene 1 in Figure \ref{fig: scene error comparison}). The blue circle shows the proximity radius of each waypoint, and the thin purple line shows the path along which the robot traveled. 
As expected, the robot using the neural ODE model in Figure \ref{fig: phoenix extrapolation figure forest node} fails to extrapolate to the unknown ice terrain, where the surface friction in Phoenix is set extremely low. The robot only reached two waypoints because it slides out of control, collides with a tree, and was unable to recover. On the other hand, the robot using the function encoder model in Figure \ref{fig: phoenix extrapolation figure forest fe} reaches all three waypoint goals without colliding with any obstacles. Over four trials, the function encoder had no collisions, while the neural ODE had 11 total. 

\begin{figure}
    \centering
    \includegraphics[width=\columnwidth]{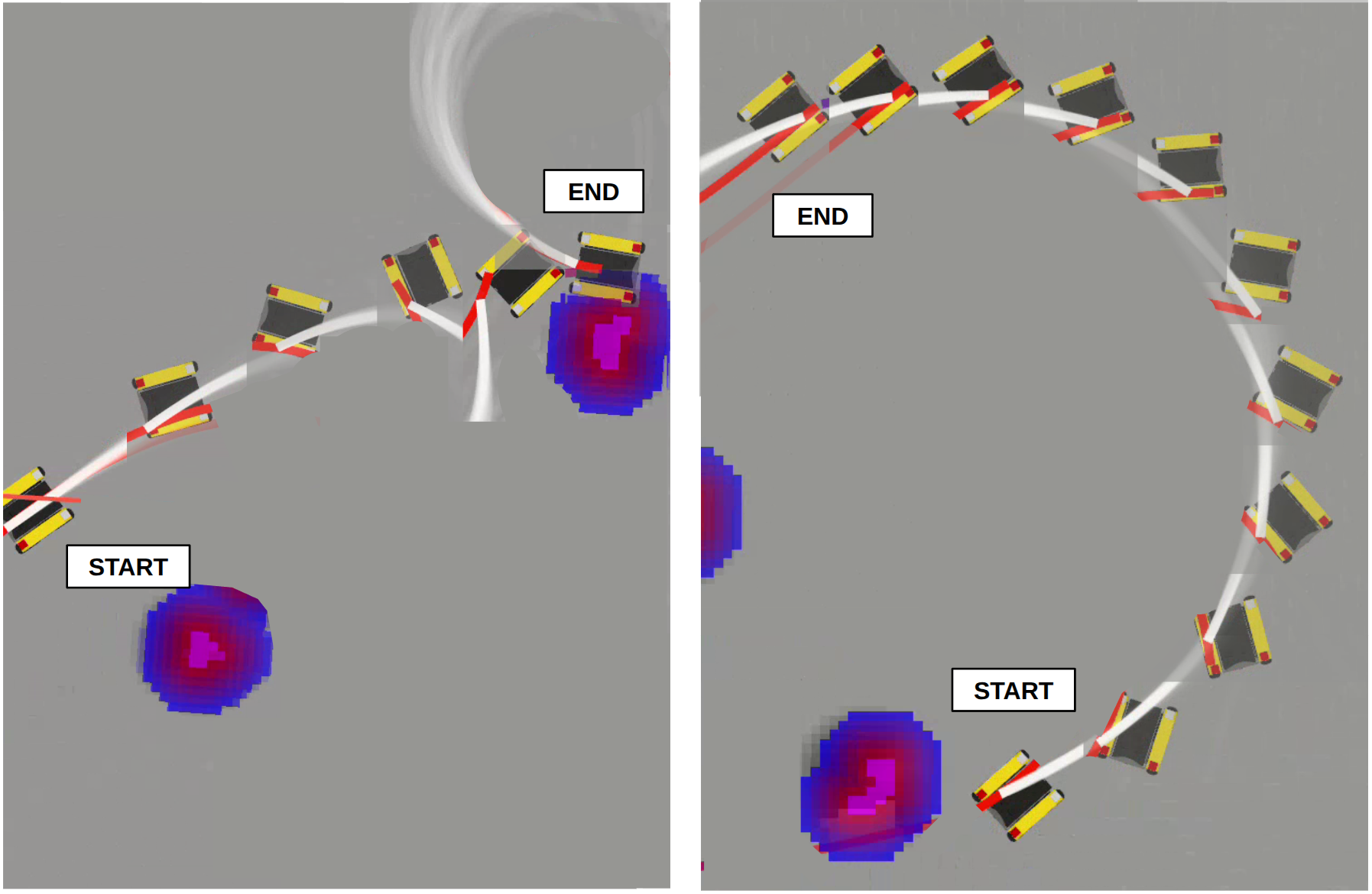}
    \caption{Timelapse illustration of model predictions. The neural ODE trajectory predictions break down during a high speed turn (left). As a result, MPPI loses control of the vehicle and collides with an obstacle. The function encoder accurately predicts the robot's trajectory during a similar maneuver (right) and maintains control.}
    \label{fig: phoenix fe node rollout comparison}
\end{figure}

Figure \ref{fig: phoenix fe node rollout comparison} illustrates each model's prediction accuracy in a timelapse. MPPI attempts to control the robot through an arc maneuver while traveling at high speeds on a slippery terrain. This is very challenging due to low friction between the tires and the ground that causes the robot to slide through turns. The function encoder better predicts how the robot slides while turning and maintains control of the vehicle. On the other hand, the neural ODE predicts with less accuracy, and MPPI loses control. 

\section{Limitations}

The main limitations of our approach are related to the data and the limitation of our validation results to a simulated robot environment.

\textit{Sensor Modality:} The current implementation relies exclusively on proprioceptive measurements (e.g.\ odometry and commanded velocities).
Consequently, adaptation is inherently reactive, meaning it occurs only after the robot has physically encountered the terrain.
This limits the ability to anticipate upcoming terrain transitions and can result in control degradation, particularly during abrupt surface changes. Nevertheless, we anticipate that our approach could likely be adapted to streaming data via recursive least squares or Bayesian approaches to update the coefficients in response to new data.

\textit{Model Expressiveness and Excitation:} The accuracy of the adapted model is constrained by the span of the learned function space.
If terrain-induced dynamics fall outside this span---e.g.\ due to highly nonlinear effects like slippage or discontinuous contacts---the model may fail to generalize, limiting performance in extreme conditions.

At runtime, the method requires only a small number of transitions (e.g.\ $\sim$100) to estimate terrain-specific coefficients via least-squares. However, this estimation assumes persistently exciting control inputs; under low-excitation conditions, such as steady, straight-line driving, the resulting regression problem may be ill-conditioned, reducing adaptation quality.

\textit{Simulation-Only Validation:} All experiments are conducted in high-fidelity simulation using a Unity-based physics engine with realistic terrain modeling. While these results are promising, real-world validation on hardware is needed to assess performance and robustness under real-world conditions.

\section{Conclusion \& Future Work} 
\label{sec:conclusion}

This work introduces a method for online adaptation of terrain-aware dynamics using function encoders, which represent robot dynamics as a linear combination of basis functions learned from diverse terrain data. 
By projecting dynamics onto a compact function space representation and estimating terrain-specific coefficients online via least squares regression, the method achieves fast and data-efficient adaptation without retraining or fine-tuning.
Integrated into a model predictive path integral (MPPI) controller, the adapted model enables accurate trajectory prediction and robust closed-loop performance on previously unseen terrain.
Experiments in a high-fidelity Unity simulation with a Clearpath Warthog robot show that our approach reduces model prediction error and results in fewer collisions compared to a neural ODE baseline, demonstrating its effectiveness for terrain-aware planning in unstructured environments.

Future work aims to incorporate visual perception into the adaptation process, enabling anticipatory, zero-shot dynamics estimation directly from visual terrain features.
Specifically, we plan to map embeddings from onboard RGB cameras to dynamics representations in the learned function space, enabling the robot to infer terrain dynamics before contact. 
We also plan to extend this work to other types of terrain variations. This work focuses on varying friction parameters, but future work seeks to evaluate our method on uneven, bumpy surfaces that are common in real-world, off-road navigation.

\section*{Acknowledgments}
This work was supported by the National Science Foundation under NSF Grant Numbers 2214939 and 2409535. Any opinions, findings and conclusions or recommendations expressed in this material are those of the authors and do not necessarily reflect the views the National Science Foundation.

This work was also supported by the Air Force Office of Scientific Research under award FA9550-19-1-0005, as well as the Office of Naval Research under award ONR N00014-20-1-2115. Any opinions, findings and conclusions or recommendations expressed in this material are those of the authors and do not necessarily reflect the views the U.S. Department of Defense.

Research reported in this paper was sponsored in part by the DEVCOM Army Research Laboratory under Cooperative Agreement W911NF23-2-0211.
The views and conclusions contained in this document are those of the authors and should not be interpreted as representing the official policies, either expressed or implied, of the DEVCOM Army Research Laboratory or the U.S. Government.
The U.S. Government is authorized to reproduce and distribute reprints for Government purposes notwithstanding any copyright notation herein.

\bibliographystyle{plainnat}
\bibliography{bibliography}

\end{document}